\documentclass[10pt,journal,cspaper,compsoc]{IEEEtran}
\ifCLASSOPTIONcompsoc
\else
\fi
\ifCLASSINFOpdf
   \usepackage[pdftex]{graphicx}
\else
\fi
\begin{document}
%
\title{Quantifying and Visualizing Vascular Branching Geometry: Normalization of Intra- and Inter-Specimen Variations}
%
%
%
%

\author{Timothy~L~Kline
\IEEEcompsocitemizethanks{\IEEEcompsocthanksitem Timothy L Kline is with the Department of Radiology,
Mayo Clinic College of Medicine,
Rochester MN, 55905 USA, E-mail: kline.timothy@mayo.edu\protect\\

}
\thanks{}}

\IEEEcompsoctitleabstractindextext{%
\begin{abstract}
Micro-CT images of the renal arteries of intact rat kidneys, which had their vasculature injected with the contrast agent polymer Microfil, were characterized. Measurement of inter-branch segment properties and the hierarchical structure of the vessel trees were computed by an automated algorithmic approach. The perfusion territories of the different kidneys, as well as the local diameters of the segmented vasculature were mapped onto the representative structures and visually explored. Various parameters were compared in order to outline key geometrical properties, properties which were shown to not have a wide range of inter-specimen variation. It is shown that the fractal scaling in non-symmetric branching reveals itself differently, than in symmetric branching (e.g., in the lung the mean bronchial diameters at each generation are closely related). Also, perfused tissue is shown to have very little inter-specimen variation and therefore could be used in future studies related to characterizing various disease states of tissues and organs based on vascular branching geometry.\end{abstract}

\begin{IEEEkeywords}
Anatomy Functional, Blood Vessel Morphology and Structure, Centerline Extraction, Micro-CT, Vascular Segmentation
\end{IEEEkeywords}}

\maketitle

\IEEEdisplaynotcompsoctitleabstractindextext

%
\IEEEpeerreviewmaketitle

%
\ifCLASSOPTIONcompsoc
  \noindent\raisebox{2\baselineskip}[0pt][0pt]%
  {\parbox{\columnwidth}{\section{Introduction}\label{sec:introduction}%
  \global\everypar=\everypar}}%
  \vspace{-1\baselineskip}\vspace{-\parskip}\par
\else
  \section{Introduction}\label{sec:introduction}\par
\fi
%

%
%
%
%
\IEEEPARstart{P}{revious}, studies have uncovered the existence of a large degree of inter- and intra- variability in the geometry of vasculature \cite{zamirBlood,labarbera,kassab}. The characterization of vascular branching geometry typically includes the measurement of inter-branch segment lengths and diameters, as well as the angle at which the vessel tree bifurcates. Based on this information, various models have been developed which give a general description for how the vascular geometry behaves \cite{yang}. In order for variations in vascular geometry to be elucidated, particularly in terms of what may occur in various states of disease, parameters with little inter-specimen variations (for specimens in the same relative state of health or disease) need to be outlined. The variation shown by previous characterizations is sufficiently great that it obscures variations resulting from various pathophysiological states. 

\section{Materials and Methods}
\label{sec:format}

\subsection{Specimens}
\label{ssec:subhead}

Previously prepared specimens \cite{langKidney} were used as the dataset in this study, and are here briefly described. Male Wistar rats (Charles River) weighing 200 g were injected with heparinized saline intravenously (1000 IU/animal) and then Microfil, a radiopaque liquid silicone polymer compound (MV-122; Flow Tech,. Inc,. Carver, MA), was injected into the abdominal aorta. The kidneys were then clamped and harvested for subsequent scanning.

\subsection{Micro-CT Imaging}
\label{ssec:subhead2}

The micro-CT imaging methods that were used to scan the specimens are discussed in more detail elsewhere \cite{jorgensen}. Briefly, the x-ray was generated with a Philips Spectroscopy tube with a molybdenum anode and Zirconium foil filter. The mean photon energy was approximately 18 keV. The scan data, once recorded, were normalized for the exposing x-ray intensity and then subjected to a modified Feldkamp cone beam reconstruction algorithm \cite{feldkamp} to generate a 3D volume data set. The 3D image consisted of up to 1024$^{3}$ cubic voxels, each $\sim$20 $\mu $m to a side, with gray scale proportional to the x-ray attenuation coefficient. The vascular structures (from the different specimens) attain approximately the same attenuation within their vascular lumens.

\subsection{Segmentation}
\label{ssec:subhead3}

The vascular trees were extracted from the gray scale image by means of a region-growing segmentation method to include those voxels, with gray scale values above above a specified threshold, that were connected to the vessel tree's root. In order to include small vessel segments, a low CT-threshold (much lower than half the maximum value found in large vessel lumens) was chosen, in order to include vessel segments that had a reduced peak gray scale value caused by the MTF (blurring) of the micro-CT system. The maximum gray scale found in large opacified lumens was $\sim$6000 [1000/cm], and a threshold of 1500 [1000/cm] was used to segment the vasculature. Fig. 1 shows an example of one of the segmented vascular trees. The region-growing method resulted in a single connected tree structure for all specimens. The kidney tissue was segmented by the same region-growing threshold method for voxels within a CT gray scale range of 600 to 1200 [1000/cm].

\begin{figure}[htb]

\begin{minipage}[b]{1.0\linewidth}
  \centering
  \centerline{\includegraphics[width=8.5cm]{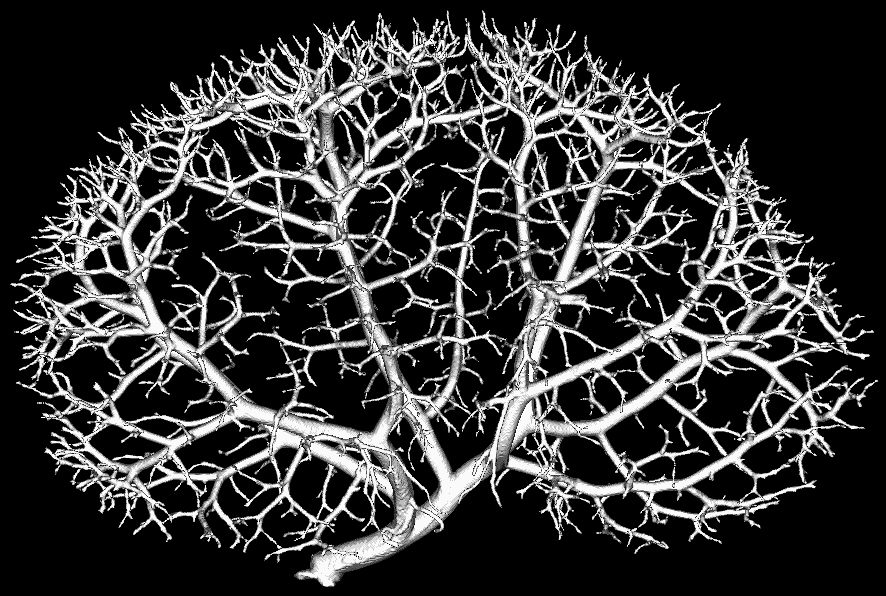}}
\end{minipage}
\caption{Volume rendered display of segmented renal artery from 3D micro-CT image data.}
\label{fig:res}
\end{figure}

\subsection{Vascular Tree Measurements}
\label{ssec:subhead3}

The 3D segmented data set was used as the input for the extraction of the vascular tree's centerline (``skeleton''). The extraction of the vascular tree centerline allows for the simplification of the complex tree structure from which the branch segment location within the tree, and its connection to the other branch segments, is conveyed. This is essentially the first step in defining the vascular tree topology. Using locations along the centerline within the segmented image, the tree's hierarchy, branch lengths, and branch diameters were determined. More details on the centerline extraction method, and subsequent measurements of inter-branch segment lengths and diameters can be found in \cite{klineABME}.

\subsection{Perfusion and Local Diameter Maps}
\label{ssec:subhead3}

The segmented vascular tree's voxels were mapped by computing the distance between a vascular tree voxel and its nearest centerline voxel. Fig. 2 shows the local diameter mapping applied to two segmented vascular trees. The segmented vascular tree mapping corresponds to the local diameter of the vasculature in each region and could visually assist in identifying stenotic regions. The kidney tissue voxels were mapped by computing the distance between each tissue voxel and its nearest vascular voxel. Shown in Fig. 3 is the perfusion map applied to one of the kidney tissue volumes, here viewed from two different directions.  The kidney tissue mapping corresponds to a regional perfusion map that gives quantitative data related to the distribution of vasculature throughout the kidney, and visually illustrates regions of reduced perfusion.

\begin{figure}[htb]

\begin{minipage}[b]{1.0\linewidth}
  \centering
  \centerline{\includegraphics[width=8.5cm]{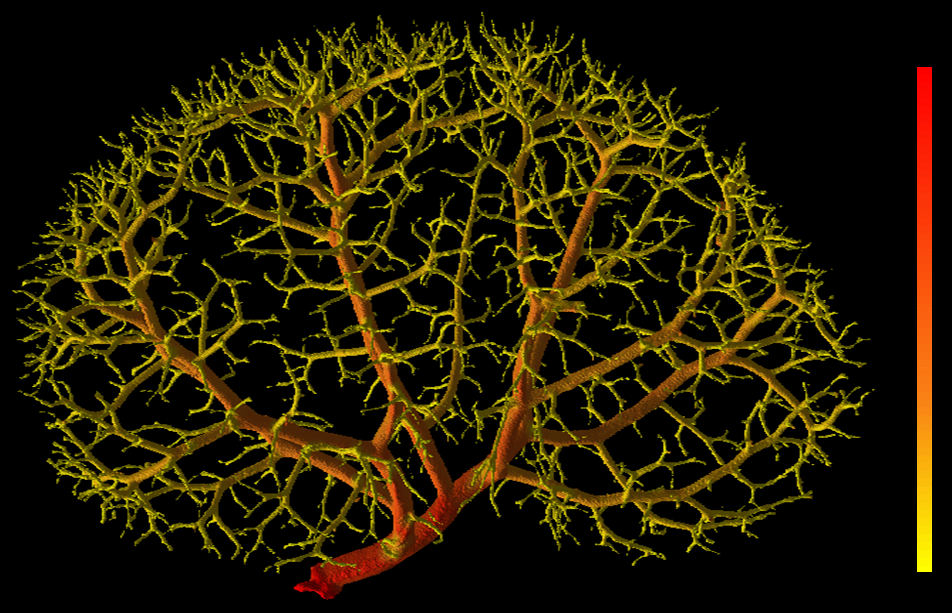}}
\end{minipage}
\caption{Local diameter mapping applied to the segmented vascular tree from Fig. 1. The regions visualized in red are larger diameter portions of the vessel tree, and the yellow colored regions represent the smallest diameter branch segments.}
\label{fig:res}
\end{figure}

\begin{figure}[htb]

\begin{minipage}[b]{1.0\linewidth}
  \centering
  \centerline{\includegraphics[width=8.5cm]{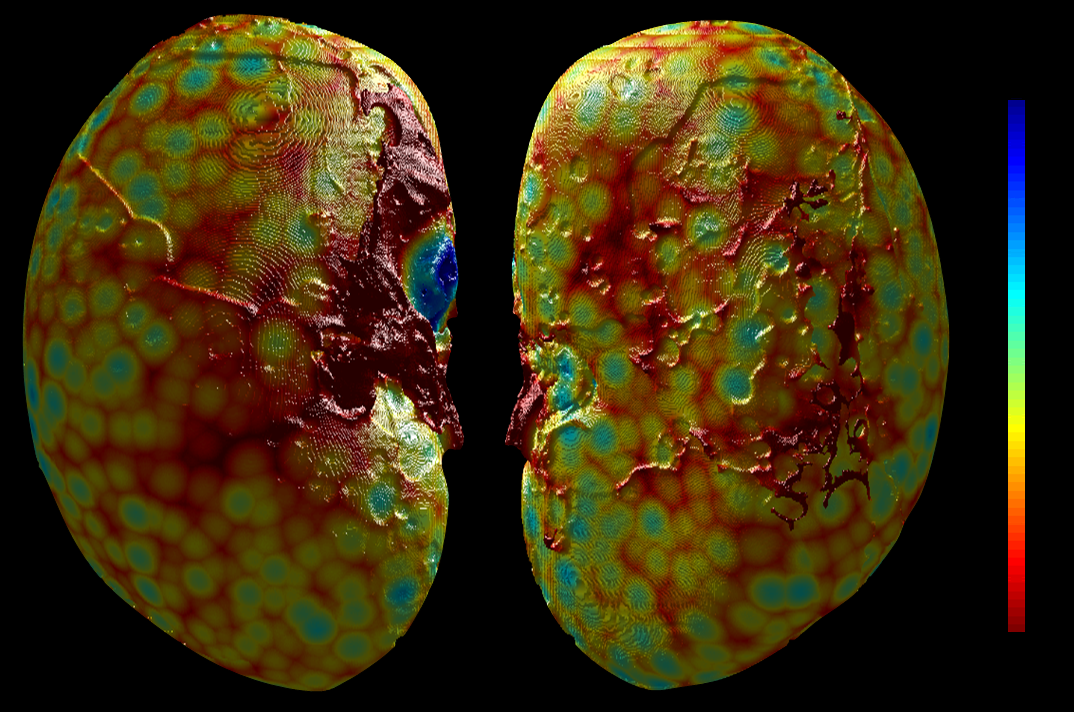}}
\end{minipage}
\caption{Perfusion map applied to one of the kidney cortex surfaces, viewed from two different directions. The tissue in the cortex regions, colored in red, are located the farthest distance from the segmented vasculature, whereas the cooler color regions are located close to the kidney vasculature. The circular regions are around superficial arterioles.}
\label{fig:res}
\end{figure}

\section{Results}
\label{sec:pagestyle}

Shown in Fig. 4 are the average diameter and standard deviation (represented as `error' bars) for the vessel segments at each generation of the tree. This is a commonly presented type of figure in studies involving the characterization of vascular branching geometry \cite{zamirBlood}. Clearly, a large range of inter-specimen variation exists, and the high degree of asymmetry in the branching geometry means that the average generation diameter is relatively meaningless. Similarly, Fig. 5 shows the number of inter-branch segments found at each generation. These are two examples of data which show a large range of variation between specimens. The large degree of inter specimen variation was also observed in the average length found at each generation (Fig. 6). 

\begin{figure}[htb]

\begin{minipage}[b]{1.0\linewidth}
  \centering
  \centerline{\includegraphics[width=8.5cm,clip=true, viewport= 0.8in 1.5in 8.5in 6.1in]{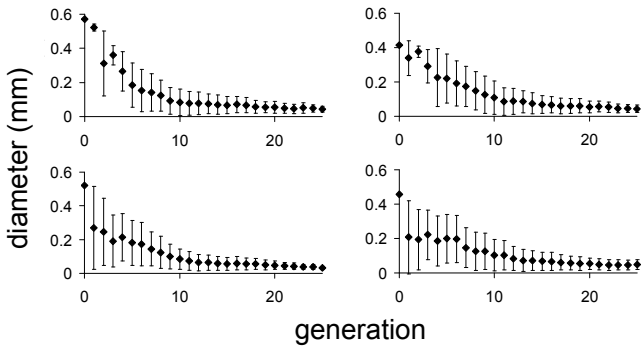}}
\end{minipage}
\caption{Average diameter and standard deviation (error bars) for the vessel segments at each generation of the tree for all four specimens. A large range of inter-specimen variation exists.}
\label{fig:res}
\end{figure}

\begin{figure}[htb]

\begin{minipage}[b]{1.0\linewidth}
  \centering
  \centerline{\includegraphics[width=8.5cm,clip=true, viewport= 1.4in 1.7in 8.7in 6.3in]{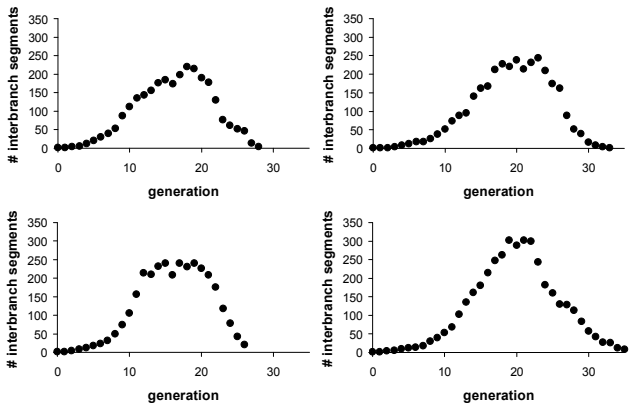}}
\end{minipage}
\caption{Number of inter-branch segments found at each generation for the four renal artery specimens. This figure, along with the previous two figures shows how many parameters of vascular branching geometry have a large range of variation between different specimens.}
\label{fig:res}
\end{figure}

\begin{figure}[htb]

\begin{minipage}[b]{1.0\linewidth}
  \centering
  \centerline{\includegraphics[width=8.5cm,clip=true, viewport= 1.4in 1.6in 8.7in 6.2in]{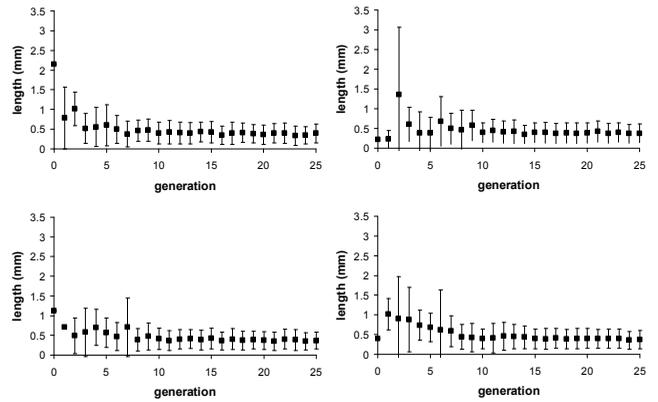}}
\end{minipage}
\caption{Average length and standard deviation (error bars) for the vessel segments at each generation of the tree for all four specimens. Like that of the average interbranch segment diameter, a large range of inter-specimen variation exists.}
\label{fig:res}
\end{figure}

\begin{figure}[htb]

\begin{minipage}[b]{1.0\linewidth}
  \centering
  \centerline{\includegraphics[width=8.5cm,clip=true, viewport= 1.4in 1.5in 8.8in 6.2in]{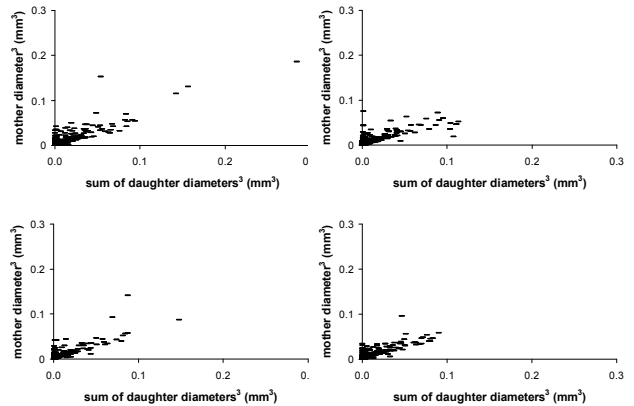}}
\end{minipage}
\caption{Murray Relation.}
\label{fig:res}
\end{figure}

Shown in Fig. 6 is the range of all specimens for the cumulative distribution of the number of segments with diameter greater than the corresponding value along the x-axis. In the intermediary region, between the first few branches supplying the kidney and the terminal arterioles, the distribution follows a power law distribution of the form

\begin{equation}
N(d) = d^{-\gamma},
\end{equation}

\noindent where $N$ is the number of inter-branch segments with diameters greater than $d$. In particular, this region follows a power-law distribution with a $\gamma = 1.5$. This was closely followed in all cases. 

The distribution of perfusion values is shown in Fig. 7. In this figure the mean and standard deviation between different specimens is depicted. Here very little variability was observed between specimens. This means that the distance of kidney tissue to the vasculature is reproducible in different specimens. This property, along with the cumulative distribution shown in Fig. 6 are examples of properties which show very little inter-specimen variation, and therefore form the basis for creating vascular models for tissue and organs in various states of health and disease.

\begin{figure}[htb]

\begin{minipage}[b]{1.0\linewidth}
  \centering
  \centerline{\includegraphics[width=8.5cm,clip=true, viewport= 2.1in 2.2in 8.3in 6.3in]{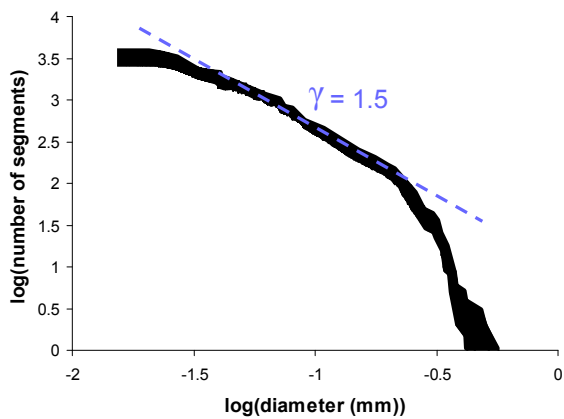}}
\end{minipage}
\caption{Cumulative distribution of the number of segments with diameter greater than the corresponding value along the x-axis. The data range is shown (i.e., all specimen data lie within the black object). Also, a range of diameters follows a power-law distribution with $\gamma = 1.5$ ( the negative slope of the dashed line regression, see Eq. 1) was observed in all cases.}
\label{fig:res}
\end{figure}

\begin{figure}[htb]

\begin{minipage}[b]{1.0\linewidth}
  \centering
  \centerline{\includegraphics[width=8.5cm,clip=true, viewport= 2.1in 2.3in 9.0in 6.3in]{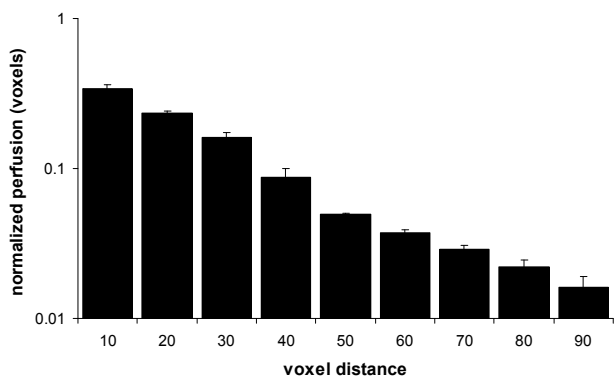}}
\end{minipage}
\caption{Distribution of perfusion values shown as the mean and standard deviation between different specimens. The y-axis is here plotted on a logarithmic scale. This property, along with the cumulative distribution shown in Fig. 6 are examples of properties which show very little inter-specimen variation, and therefore form the basis for creating vascular models for the vasculature of various tissues and organs in different states of health and disease.}
\label{fig:res}
\end{figure}
\section{Discussion}
\label{sec:typestyle}

The study of vascular branching geometry is an area of anatomy which has yet to reach the same level of accuracy and understanding as other areas. The main reason for this has been the methodological difficulties in obtaining accurate data, and in building a cumulative understanding based on these data.

Micro-CT, which is able to quantify anatomical information in three-dimensions non-destructively at a very high spatial resolution, is a great tool for studying questions related to microcirculation. However, using micro-CT imaging to quantify vascular branching geometry requires solutions to many technical problems, such as problems associated with blurring due to the modulation transfer function (MTF) of the scanner's imaging system \cite{kalsho}, the partial volume effect \cite{glover}, differences in gray scale of the micro-CT image depending on the tomographic approach used \cite{herman}, and identifying vasculature within the image (i.e., segmentation) \cite{leeSegmentation}. In this study we have used a previously validated method \cite{klineABME} (which accounts for many of these technological problems) to gather quantitative data of rat renal arteries. 

To develop models of vascular systems in various states of health and disease requires the elucidation of key geometrical properties that are retained between different specimens in the same pathophysiological state. It was shown that the power law distribution of the cumulative diameter of inter-branch segments (which considers the number and size of inter-branch segments, as opposed to the hierarchical nature of the tree) appears to align inter-branch segment data more closely with their role in microcirculation. Thus, the fractal scaling in non-symmetric branching reveals itself differently, than in, for example, the lung \cite{goldberger} where mean bronchial diameters at each generation are much more closely related. This, along with the distribution of perfused tissue (as it relates to the vasculature), were shown to have very little inter-specimen variation, and therefore could be used in future studies related to characterizing various disease states of tissues and organs based on the local vascular branching geometry. It remains to be shown how early pathophysiological status is conveyed by the branching geometry variables discussed in this work.  
\ifCLASSOPTIONcaptionsoff
  \newpage
\fi



%
\bibliographystyle{IEEEbib}
\bibliography{P:/images/kline_t/Work2/kline_t/references/myrefs}
%






\end{document}